\documentclass[10pt,twocolumn,letterpaper]{article}

\usepackage[pagenumbers]{cvpr} % To force page numbers, e.g. for an arXiv version

\usepackage[dvipsnames]{xcolor}

\usepackage[accsupp]{axessibility} % Improves PDF readability for those with disabilities.
\usepackage{tabularx}
\newcommand{\ind}{\perp\!\!\!\!\perp}
\DeclareMathOperator*{\argmin}{argmin}   

\definecolor{cvprblue}{rgb}{0.21,0.49,0.74}
\usepackage[pagebackref,breaklinks,colorlinks,citecolor=cvprblue]{hyperref}

\def\paperID{3}
\def\confName{CVPR}
\def\confYear{2024}

\title{Enforcing Conditional Independence for \\ Fair Representation Learning and Causal Image Generation}

\author{
Jensen Hwa\textsuperscript{1},
Qingyu Zhao\textsuperscript{2},
Aditya Lahiri\textsuperscript{3},
Adnan Masood\textsuperscript{4},
Babak Salimi\textsuperscript{3},
Ehsan Adeli\textsuperscript{1}\\
\textsuperscript{1}Stanford University \quad
\textsuperscript{2}Weill Cornell Medicine \quad
\textsuperscript{3}University of California San Diego \quad
\textsuperscript{4}UST \\
{\tt\small jphwa@cs.stanford.edu, eadeli@stanford.edu}
}

\begin{document}
\maketitle
\begin{abstract}
Conditional independence (CI) constraints are critical for defining and evaluating fairness in machine learning, as well as for learning unconfounded or causal representations. Traditional methods for ensuring fairness either blindly learn invariant features with respect to a protected variable (\eg, race when classifying sex from face images) or enforce CI relative to the protected attribute only on the model output (\eg, the sex label). Neither of these methods are effective in enforcing CI in high-dimensional feature spaces. In this paper, we focus on a nascent approach characterizing the CI constraint in terms of two Jensen-Shannon divergence terms, and we extend it to high-dimensional feature spaces using a novel dynamic sampling strategy. In doing so, we introduce a new training paradigm that can be applied to any encoder architecture. We are able to enforce conditional independence of the diffusion autoencoder latent representation with respect to any protected attribute under the equalized odds constraint and show that this approach enables causal image generation with controllable latent spaces. Our experimental results demonstrate that our approach can achieve high accuracy on downstream tasks while upholding equality of odds.
\end{abstract}    
\section{Introduction}
\label{sec:intro}

\begin{figure}[t!]
    \centering
    \includegraphics[width=\linewidth]{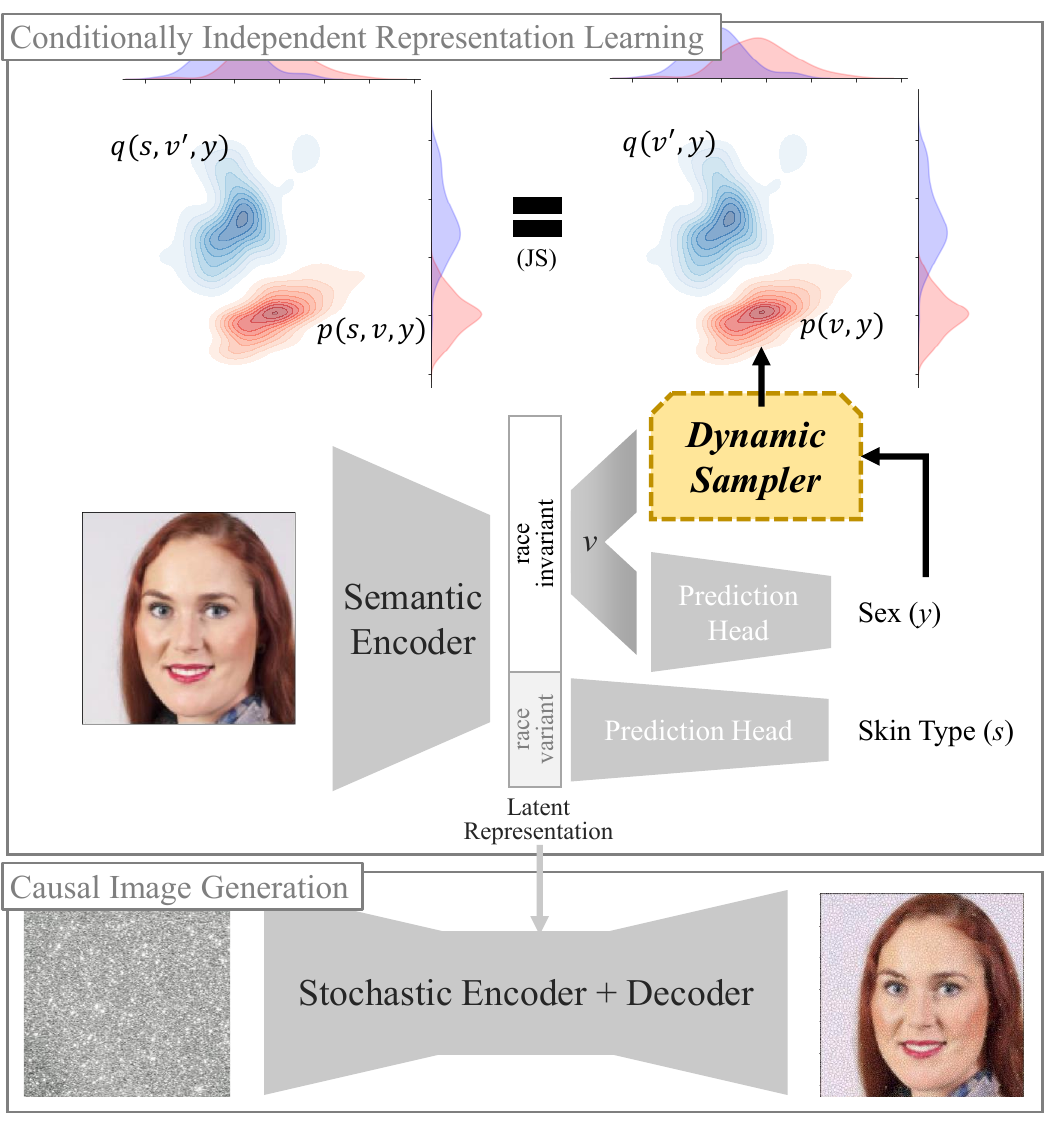}
    \caption{We propose a new way to ensure fairness in downstream tasks by enforcing conditional independence constraints on the latent representation. This is achieved by minimizing the Jensen-Shannon divergence (JS) between distributions obtained using a novel dynamic sampling technique. In the setting shown here, we apply our technique to the diffusion autoencoder's semantic representation to disentangle the sensitive attribute of skin type (a proxy variable for race) and perform causal image generation.}
    \label{fig:fig1}
\end{figure}

Fairness in machine learning and computer vision is an increasingly important topic with growing applications to everyday life. The literature includes several methods for learning fair and unconfounded models based on domain adversarial invariant learning \cite{kim2019, zhao2020training, zhang2018mitigating, madras2018learning, akuzawa2019adversarial}, statistical methods \cite{squires2020, lu2021metadata}, and information theory \cite{moyer2018}. 

In the context of predictive modeling, algorithmic fairness aims to learn models that are unbiased towards protected subgroups by ensuring that the output label $\hat{y}$ is 
{\em invariant} to a sensitive attribute $s$. The existing plethora of fairness definitions typically captures this invariance in terms of a {\em conditional independence (CI)} (see~\citeauthor{mehrabi2021survey} for a survey). For instance, the equalized odds criterion~\cite{hardt2016equality} requires prediction output $\hat{y}$ to be independent of the sensitive attribute $s$ conditioned on the true class label $y$.
We note that the conditional independence formulation $s \ind \hat{y} \mid y$ is widely recognized as a more precise indicator of fairness compared with the marginal independence formulation $s \ind \hat{y}$, due to the fact that some correlations between sensitive attributes and the target variable may be benign. CI plays a crucial role in capturing the true causal relationships between the sensitive attributes and the target variable, ensuring that fairness is based on genuine causes of disparities rather than arbitrary associations~\cite{kusner2017counterfactual,salimi2019interventional}.

However, most existing work in fair machine learning typically focuses on enforcing independence on low-dimensional features in cases where the conditioning set is either empty or merely a low-dimensional categorical variable \cite{salimi2019interventional, salimi2020, pmlr-v161-ahuja21a, zhao2020training, agrawal2021}. This restricts the learning of fair representations to limited settings involving only binary attributes in tabular datasets \cite{zhao2019conditional}. These methods cannot directly translate to high-dimensional spaces, such as the latent representations of large generative models.

Recently, \citeauthor{pmlr-v161-ahuja21a} developed a differentiable framework for enforcing CI in high-dimensional and continuous feature spaces by using a GAN-based approach in the context of data generation. 
To  enforce $s  \ind \hat{y} \mid y$, they minimize the Jensen-Shannon (JS) divergence between the joint distribution $p(s,\hat y,y)$ and  an auxiliary distribution $q(s,y',y)$ while the joint marginal $q(s,y)$ is distributed identically to $p(s,y)$ and $q(y' \mid y)$ is independent of $s$. Although this formulation enforces CI, it can only sample the joint probability distributions in the label space $y$ and does not necessarily guarantee the learning of fair representations, \ie, only the last network layer before predicting $y$ removes the dependence while the rest of the network (including latent representations) remain biased or confounded.

Our approach introduces the concept of CI into the latent space by using a \textit{dynamic sampler} to form the joint distributions with respect to the learned representations. This yields a computationally rigorous approach to fair and causal representation learning that exhibits a level of versatility not found in existing methods: We are able to control not just which features are encoded, but also where within the representation they appear. With this, we gain improved classification performance and, in the case of generative models such as the diffusion autoencoder (DiffAE), the ability to maintain fine-grained control over generated content. As a result, we can not only protect downstream tasks from bias against specific attributes, but also generate feature-invariant images corresponding to the enforced schema, allowing for a richer understanding of a model's representations.

\section{Related Work}
\label{sec:relatedwork}
\begin{figure}
    \centering
    \includegraphics[width=\linewidth]{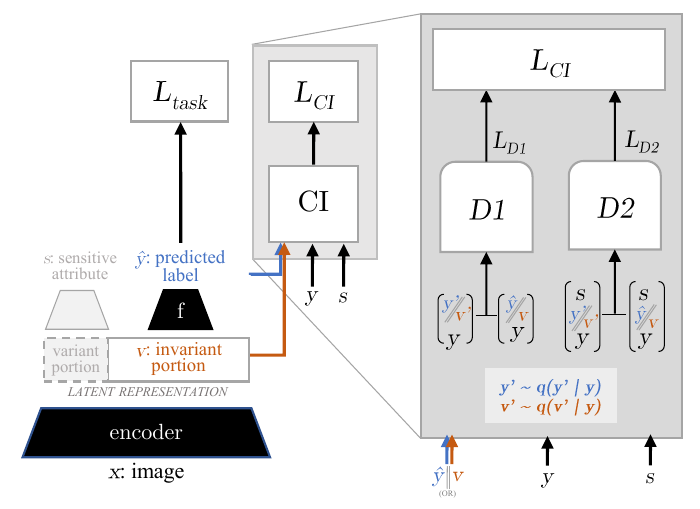}
    \caption{High-level view of our architecture. We introduce two variants of a conditional independence enforcer that can be added to any off-the-shelf encoder.}
    \label{fig:complete_architecture}
\end{figure}

\paragraph{Invariant Representation Learning}
In the wake of increasingly biased large-scale models \cite{hardt2016equality, adeli2021, madras2018learning, wang2019balanced}, the learning of fair and unconfounded representations has taken on outsized importance. Methods based on domain adversarial learning are increasingly popular tools of choice in reducing bias and confounding effects \cite{zhang2018mitigating, adeli2021, zhao2020training, li2018domain, zhang2022enhanced}. Among these, 
\citeauthor{johndrow2019algorithm} and 
\citeauthor{tan2020learning} proposed methods for invariant feature learning and 
\citeauthor{zhao2020training} 
presented models that minimize statistical mean dependence using a correlation-based adversarial loss function. Recently, \citeauthor{pirhadi2024otclean} introduced a data cleaning method that enforces the conditional independence constraint for tabular data using optimal transport.

Other types of methods incorporating statistical operations have also been explored in the literature, such as using multivariate regression analysis \cite{mcnamee2005regression} with general linear models \cite{zhao2019confounder}. Other approaches enforce fairness through post-processing steps on unfair trained classification models \cite{feldman2015computational, hardt2016equality, zhao2019confounder}. But since the training and fairness enforcement steps are conducted separately, these algorithms often lead to suboptimal fairness and accuracy trade-offs \cite{woodworth2017learning}. Recently, using traditional statistical methods, \citeauthor{lu2021metadata} proposed a normalization layer that corrects the feature distributions with respect to labeled sensitive attributes. Their approach, known as metadata normalization, entailed a simple layer that could plug into any end-to-end model to protect the models against bias. \citeauthor{vento2022penalty} then extended this work by turning the closed-form normalization operation into a network optimizable step. This is an active research field and many other approaches are being studied.

\paragraph{Conditional Independence for Algorithmic Fairness}
Consider a classifier or regression function $f(x)$ with output $\hat{y}$ and a {\em protected attribute} $s$, \eg, sex or race.
Algorithmic fairness aims to learn prediction functions or transformations that make the final outcome invariant or insensitive to protected attributes.
Several widely used {\em associational} and {\em causal} notions of fairness, including the equalized odds criterion, can be attained by means of independence \cite{salimi2020}.

Rather than attempting to transform already trained unfair models to fair ones in a post-hoc manner, it is critical to enforce fairness constraints from the beginning of the model training pipeline \cite{islam2022through}. Apart from preventing the propagation of bias to downstream modeling tasks, this approach of fairness from a data management standpoint also leads to more robust and significant fairness measures by ensuring that data sources, transformations, and other training assumptions are sound \cite{salimi2019interventional}.

\paragraph{Representation Learning with Diffusion Models}
Diffusion-based models learn to generate images by a denoising process: Random Gaussian noise is added to input images, and the model learns how to reverse this process to hallucinate new images from random noise. This family of models has proven to offer remarkable quality in image generation, promising to replace generative adversarial networks (GANs) \cite{goodfellow2020generative} as the dominant architectural paradigm. Chief among them is the diffusion autoencoder \cite{preechakul2022diffusion}, which encodes an image into a two-part latent subcode, capturing a stochastic representation via the aforementioned denoising approach and conditioning this process upon a semantic representation learned by a CNN. Interpolation of the resulting subcode results in smooth and meaningful changes in the decoded image. The semantic separation  capability of DiffAEs has enabled various applications, such as attribute manipulation and various low-shot or zero-shot downstream applications \cite{sinha2021d2c,giannone2022few,guan2020large}. In this work, we further enable DiffAEs to learn unbiased, causal, and conditionally independent latent subcode representations.
\section{Method}
\label{sec:method}

Let $\{x,y,s\}$ be a training sample in the dataset, where $x$ is an input image, $y$ is the target prediction label, and $s$ is a protected sensitive attribute. We aim to learn an encoder network $g_{\theta}(x)=v$ resulting in a latent representation $v$, from which a prediction network $f_\phi(v)=\hat y$ produces the final predicted label $\hat y$. With a slight abuse of notation, we use lower-case letters to also denote random variables. Ultimately, we aim to train a model that achieves high accuracy in predicting label $y$ while ensuring equalized odds: 
\begin{equation}
    \label{eqn:objective}
    \begin{split}
    \argmin_{\theta,\phi}~-\frac{1}{n}\sum_{i=1}^n y_i\log(\hat y_i)&+(1-y_i)\log(1-\hat y_i) \\
    \text{s.t.}\quad s\ind \hat y~|~y.
    \end{split}
\end{equation}

We discuss two possible ways of enforcing the equalized odds fairness constraint: First, we enforce conditional independence with respect to the label $y$, \ie, $s \ind \hat y  ~|~ y$. As discussed in \citeauthor{pmlr-v161-ahuja21a}, this can be seen as a means to enforce equalized odds through CI. However, we claim that enforcing the independence constraint with respect to only the label is limited in its ability to enforce the equalized odds fairness constraint effectively. We therefore additionally propose to enforce the conditional independence with respect to the more information-rich latent representation $v$, or $s \ind v  ~|~ y$. We can then use this learned fair latent representation in a downstream prediction or generation task.

\subsection{Enforcing CI with respect to Label}
\label{sec:label_space_ci}
When the label $y$ is highly correlated with the protected attribute $s$ in the training data, a basic aim for a fair machine learning model is to ensure conditional independence of the predicted $\hat y$ from $s$, \ie, $s \ind \hat y  ~|~ y$. 

Assuming that the joint distribution $p(s,\hat y,y)$ is well-defined with respect to Lebesgue measure $\mu$, previous work~\cite{pmlr-v161-ahuja21a} demonstrates that the above conditional independence can be defined with respect to the Jensen-Shannon divergence (JS) between $p(s,\hat y,y)$ and an auxiliary distribution $q(s,y',y)$, where the joint marginal $q(s,y)$ is distributed identically to $p(s,y)$, and $q(y'|y)$ is independent of $s$. In other words, when $q(s,y',y)=p(s,y)q(y'|y)$, the conditional independence constraint $s \ind \hat y  ~|~ y$ is equivalent to enforcing
\begin{equation}    
\label{eqn:jsd}
\text{JS}(p(\hat y, y), q(y', y)) = \text{JS}(p(s,\hat y, y), q(s, y', y)).
\end{equation}
Accordingly, we can rewrite~\eqref{eqn:objective} in terms of JS as
\begin{equation}
    \label{eqn:objective-jsd}
    \begin{split}
    \argmin_{\theta,\phi}~&-\frac{1}{n}\sum_{i=1}^n y_i\log(\hat y_i)+(1-y_i)\log(1-\hat y_i)\\
    \text{s.t.} \quad & \big(\text{JS}(p(\hat y, y), q(y', y)) -\\ &
    \text{JS}(p(s,\hat y, y), q(s, y', y))
    \leq \delta \big)
    \end{split}
\end{equation}
for some $\delta>0$ sufficiently small.

With the help of a conditional sampler $q(y' ~|~y)$, the JS constraint of \eqref{eqn:objective-jsd} can be satisfied by a general GAN architecture with two discriminators (Fig. \ref{fig:complete_architecture}) \cite{pmlr-v161-ahuja21a}, \ie, by minimizing 
\begin{equation}
L_\text{CI} = (L_\text{D1} - L_\text{D2})^2
\label{eqn:ci}
\end{equation}   
where
\begin{align*}
    L_\text{D1} &= \mathbb{E}[\text{log}(1-D_1(y',y)) + \mathbb{E}_{y,s}[\log D_1(\hat y, y)]  \quad\text{and} \\
    L_\text{D2} &= \mathbb{E}[\text{log}(1-D_2(y',s,y)] + \mathbb{E}_{y,s, y'}[\log D_2(\hat y, s, y)].
\end{align*}

The key advantage of using \eqref{eqn:ci} to derive conditional independence is that $q(y'|~y)$ does not have to be a perfect sampler that exactly matches $p(\hat y~|~y)$. The only sufficient condition is that $q(y'|~y)$ shares overlapping support with the original distribution $p(\hat y~|~y)$. This gives us many flexible ways to construct $q$, such as using a uniform distribution.

Specifically, in a supervised learning setting, where $y$ is known at training time, we can set $q$ using a uniform distribution over all raw model outputs that correspond to $y$. The encoder and prediction networks can then be learned by optimizing the loss function
\begin{equation}   
L_{\text{Enc}} = \lambda L_{\text{CI}} + L_{\text{task}},
\label{eqn:enc_loss}
\end{equation}
where $\lambda$ is a hyperparameter and $L_{\text{task}}$ is the loss function for the prediction task. The discriminators can be trained adversarially to minimize the composite loss function: 
\begin{equation}   
L_{\text{D}} = L_{\text{D1}} + L_{\text{D2}}.
\label{eqn:d_loss}
\end{equation}

\subsection{Enforcing CI with respect to Latent Representation}
\label{sec:resampling}
Enforcing conditional independence only with respect to $\hat y$ does not necessarily prevent the model from learning biased information from $s$ in its representations. In this paper, we propose to enforce conditional independence with respect to the latent representation $v$ in a way that most directly impacts the learnable parameters of the encoder. We achieve this by replacing the predicted label $\hat y$ with the latent vector $v$ within our conditional independence loss described above, \ie, $s \ind v  ~|~ y$. Similar to \eqref{eqn:jsd}, our Jensen-Shannon divergence constraint becomes 
\begin{equation}
\label{eqn:jsd_latent}\text{JS}(p(v, y), q(v', y)) = \text{JS}(p(s,v, y), q(s, v', y)).
\end{equation}
As before, this requires a conditional sampler $q(v'|y$). However, one problem that arises with this approach is that the conditional sampler is no longer well-defined at training time. Since the distribution and support of $p(v|y)$ is learned by the model on the fly, we cannot construct $q(v'|y)$ a priori. 

To resolve this challenge, we implement the imperfect conditional sampler $q$ via a novel \textbf{dynamic sampling} procedure. Specifically, to sample $v' \sim q(v'|y)$, we sample with replacement from the latent vectors associated with the given $y$ within the same training batch. Similar to the bootstrapping procedure, our dynamic sampling can approximate the empirical distribution function $p(v|y)$, allowing the discriminators to keep up with and robustly train the encoder as the latent space is learned. Note that, other than the difference in the sampler, the objective function of \eqref{eqn:ci} and the GAN architecture in Fig. \ref{fig:complete_architecture} can be translated here by replacing $\{\hat y, y'\}$ with $\{v, v'\}$.

\subsection{Disentangling DiffAE Latent Representation}

The above construction of $q$ provides us with a powerful technique that can be applied to any encoder architecture. We demonstrate this by using the state-of-the-art diffusion autoencoder model \cite{preechakul2022diffusion}, which encodes images using a denoising diffusion process \cite{sinha2021d2c} conditioned upon a semantically meaningful representation inferred by a CNN encoder. An image, therefore, is encoded by two representations: a semantic subcode learned by the CNN, along with a stochastic subcode that, given the semantic subcode, denoises into the original image. The model learns to compress the most common high-level features into the semantic subcode and delegate the remaining, highly variable features to the stochastic representation. When applied to human face images, this yields a clean separation between anatomical features such as sex, facial structure, and skin type in the semantic subcode, and more transient qualities like pose, hairstyle, and expression in the stochastic subcode.

By enforcing conditional independence with respect to the semantic subcode of the diffusion autoencoder, we produce a latent representation that is invariant to a protected attribute of choice. The model adapts by encoding any facial features associated with the protected attribute in the stochastic subcode instead. When such facial features include the high-level attributes described earlier, this can unduly affect the natural dichotomy between the semantic and stochastic subcodes, restricting the expressive power of the semantic subcode and reducing the efficiency of the denoising process used for image generation. Empirically, this results in blurred and unrealistic generated images.

To overcome this, we enforce conditional independence with respect to only a portion of the semantic subcode. To further encourage the model to use the remaining portion to encode the facial features related to the sensitive attribute, we add a prediction head on the remaining portion to estimate the sensitive attribute $s$. We optimize this new prediction head alongside the encoder; instead of \eqref{eqn:enc_loss}, we have
\begin{equation}
    L_{\text{Enc}} = \lambda_1 L_{\text{CI}} + \lambda_2~\text{BCE}(\hat s, s) + L_{\text{task}}.
\end{equation}

In this way, the semantic subcode is apportioned into two parts: one variant to the sensitive attribute, and the other invariant. Although we only demonstrate this capability on tasks involving a single sensitive attribute, this approach can be easily extended to disentangle multiple attributes within the semantic subcode.
With this formulation for causal disentanglement of the latent space, we show that the specific subcode in which a given facial attribute is represented can be methodically adjusted (\eg, to change the race of the generated image), all while preserving the model's ability to accurately reconstruct other characteristics of the input image.

\section{Experiments and Results}

We apply our architecture to two different settings, each with a unique dataset, model, and confounding variable.
In each case, we analyze the effectiveness of applying CI in the latent space as opposed to the label space and also compare with several baselines. Experiments were performed using a P100 GPU, with the exception of the diffusion autoencoder experiments which required four A100 GPUs.

\paragraph{Metrics}
We report prediction accuracies for each value of the protected attribute when $s$ is discrete, along with the balanced accuracy (bAcc) to account for class imbalances. To evaluate invariance to the sensitive attribute, we use the squared distance correlation (dcor$^2$) \cite{szekely2007measuring} to quantitatively measure the correlation between the protected variable and the learned features. When the protected variable is discrete, we also use the equality of opportunity (EO) independence metric, which measures the average gap in true positive rates for different possible values of the protected variable.
\begin{figure}[b]
    \centering
    \includegraphics[width=\linewidth]{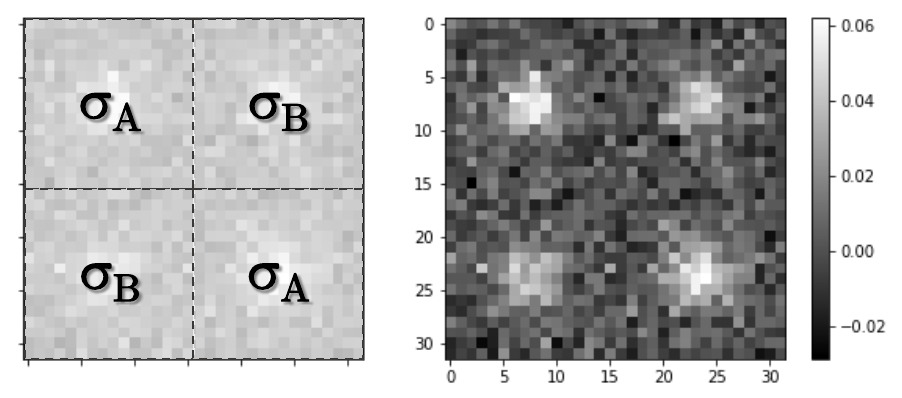}
    \caption{Synthetic data format and sample. The diagonal kernels are controlled by $\sigma_A$, while the off-diagonals are controlled by $\sigma_B$.}\vspace{-3pt}
    \label{fig:synthetic}
\end{figure}

\subsection{Synthetic Experiments}

As an initial proof of concept, we train a convolutional neural network on synthetically generated image data. Each image in this dataset is of size $32 \times 32$ and composed of four Gaussian kernels, as shown in Fig. \ref{fig:synthetic}. The diagonal kernels are of equal intensity $\sigma_A$, and the off-diagonal kernels are of equal intensity $\sigma_B$. For half of the images, we sample $\sigma_A, \sigma_B \sim \mathcal{U}(1, 4)$ and assign the label $Y = 0$. For the other half, we sample $\sigma_A, \sigma_B \sim \mathcal{U}(3, 7)$ and assign $Y = 1$. We then denote $\sigma_B$ as the protected attribute and analyze the model's ability to predict an image's label based on the diagonal kernels controlled by $\sigma_A$ while ignoring the off-diagonal kernels controlled by $\sigma_B$. We do this by enforcing the conditional independence condition separately with respect to the label space ($\sigma_B  \ind \hat Y  ~|~ Y$) and the latent space ($\sigma_B  \ind V  ~|~ Y$).

Whereas an unconstrained model may achieve maximum theoretical accuracy of $1-\frac{1}{2}\left(\frac{1}{3}\right)^2=0.94$, correctly predicting all cases except half of those where $\sigma_A, \sigma_B \in (3,4)$, a fair classifier using only the information represented by $\sigma_A$ can attain a theoretical accuracy of $1-\frac{1}{2}\left(\frac{1}{3}\right)=0.83$.

Our encoder is a simple CNN comprised of a convolutional layer, ReLU activation, max pooling, convolution, ReLU, and two linear layers, applied in that order. The first linear layer has an output of length 10, which we define as the latent space $v$, while the second linear layer converts this latent space into the single output logit corresponding to $y$.

In Table \ref{tab:synthetic-results}, we report metrics after training each model using a batch size of 512 and a learning rate of $1\mathrm{e}{-4}$. We select the $\lambda$ hyperparameter by 5-fold cross-validation and report metrics based on a separate test set. 

\begin{figure}[t]
  \centering
  \includegraphics[width=\linewidth]{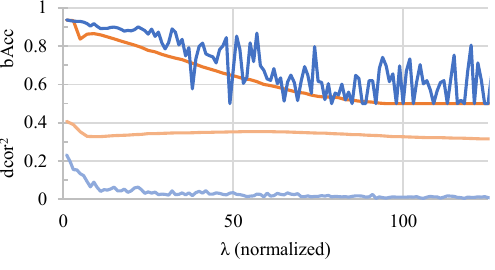}
   \caption{Fairness and accuracy metrics versus conditional independence strength $\lambda$. \textcolor{orange}{Orange} lines correspond to the \textcolor{orange}{$y$-space CI-CNN}, and \textcolor{blue}{blue} lines to the \textcolor{blue}{$v$-space CI-CNN}.}
   \label{fig:tradeoff}
   \vspace{-1em}
\end{figure}

Fig. \ref{fig:tradeoff} demonstrates the tradeoff of fairness versus accuracy and dcor$^2$ for both model variants. The latent space variant introduces some instability for larger $\lambda$, but achieves increased accuracy alongside decreased dcor$^2$. While some instability is present for large $\lambda$, this largely does not appear until after dcor$^2$ has converged.

\begin{figure*}
    \centering
    \includegraphics[width=\linewidth]{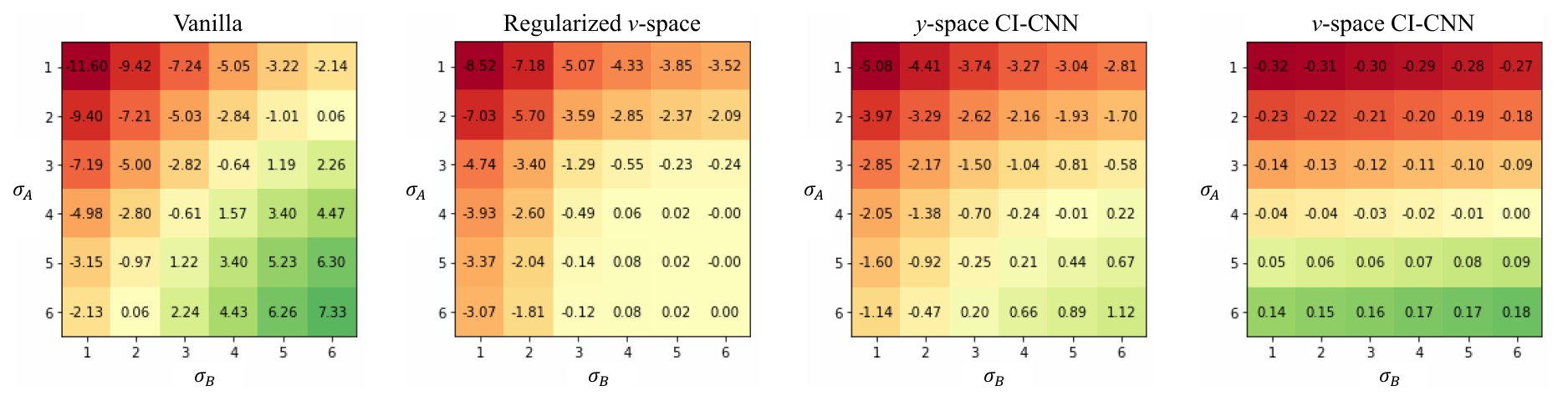}
    \caption{Unnormalized model predictions on synthetic data for given $\sigma_A,\sigma_B$. Negative values imply a prediction of $\hat y = 0$, while positive values correspond to $\hat y = 1$.}
    \label{fig:syntheticlogits}
\end{figure*}

In Fig. \ref{fig:syntheticlogits}, we show each model's unnormalized output for each integer combination of $\sigma_A$ and $\sigma_B$. Enforcing CI in the latent space produces outputs that remain nearly unaffected by changes in the protected attribute $\sigma_B$, as evidenced by near-constant values within each row.

\begin{figure}[t]
    \centering
    \includegraphics[width=\linewidth]{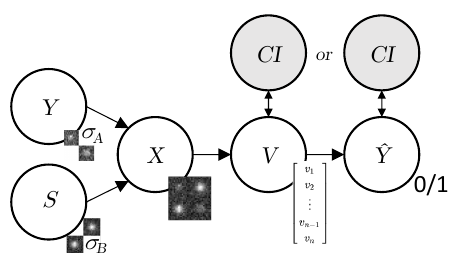}
    \caption{Synthetic data causal diagram. We apply the $L_{CI}$ component to either the latent vector space (V) or the label space (Y).}\vspace{-3pt}
    \label{fig:synthetic-diagram}
\end{figure}

\begin{table}
\caption{\label{tab:synthetic-results}Synthetic data experiment results. Balanced accuracy (bAcc) closer to $0.83$ is better, as is lower dcor$^2$.
}
\begin{tabularx}{\linewidth}{Xll}
\textbf{Model}            & \textbf{bAcc} & \textbf{dcor}$^2$  \\ \hline
Vanilla CNN               & 0.94          & 0.432              \\
Regularized $v$-space CNN & 0.80          & 0.382              \\
$y$-space CI-CNN          & 0.87          & 0.298              \\
$v$-space CI-CNN          & \textbf{0.84} & \textbf{0.055}     \\ \hline
\end{tabularx}
\end{table}

\paragraph{Model B1: Vanilla CNN} As a baseline, we train the standalone CNN without any conditional independence enforcing mechanism, using only a binary cross entropy loss function. This model relies heavily on the protected attribute $\sigma_B$, as shown by a high bAcc and $dcor^2$.

\paragraph{Model B2: Regularized $v$-space CNN}\label{para:synthetic_b2} To evaluate the efficacy of our conditional independence enforcement, we establish a baseline by embedding the equalized odds condition directly into the loss function. Defining $R_0$ and $R_1$ to be the absolute difference between the true positive rate and false positive rate for $Y=0$ and $Y=1$ respectively, the loss function becomes
\begin{equation}
    L = \lambda (R_0+R_1) + \text{BCE}(\hat y, y).
\end{equation}
We tune the $\lambda$ parameter over the validation set to target the theoretical maximum accuracy under conditional independence. This regularizer appears to have an overbearing effect on the model, and despite the reduced dcor$^2$, it struggles to learn a meaningful representation.

\paragraph{Model 1: $y$-space CI-CNN} We use the output of the final linear layer as input to our conditional independence enforcing loss component and apply binary cross entropy loss:
\begin{equation}
    L_{\text{Enc}} = \lambda L_{\text{CI}} + \text{BCE}(\hat y, y),
\end{equation}
where, as in \eqref{eqn:enc_loss}, $\lambda$ controls the strength of conditional independence enforcement and $L_{CI}$ is the loss from the dual discriminators. We find that the model learns a representation somewhat uncorrelated to $\sigma_B$, but only in an unstable manner without much improvement upon the regularized baseline.
\paragraph{Model 2: $v$-space CI-CNN} In our second iteration, we insert the conditional independence component closer to the latent space of the model, before the final linear layer, and employ the dynamic sampling procedure to simulate the conditional latent space distribution. The resulting model exhibits a much smaller confounding effect, achieving an accuracy closest to the theoretical maximum under CI and the lowest correlation between the protected and target variables. This demonstrates the effectiveness of our dynamic resampling procedure, which makes practical the enforcement of conditional independence with respect to the latent space and is one of the key contributions of this paper.

\subsection{Face Image Experiments}

\begin{table*}
\caption{\label{tab:face-results}Classification of sex from Gender Shades dataset facial images, with mean and standard deviation across five runs.
}
\centering
\begin{tabularx}{\linewidth}{Xc|cccccc|cc}
\textbf{Model}                          & \textbf{bAcc (\%)}     & I                      & II                     & III                    & IV                     & V                      & VI                     & \textbf{EO (\%)}      & \textbf{dcor${}^2$}     \\ \toprule
Kim \etal~\cite{kim2019}          & 88.8 $\pm$ 1.3 & 87.8                   & 93.0                   & 93.6                   & 91.7                   & 87.4                   & 86.2                   & 8.4 $\pm$ 3.6 & \textbf{0.022} $\pm$ 0.012          \\
Multi-task \cite{lu_multitask2017} & 93.9 $\pm$ 0.4 & 92.7                   & 95.6                   & 93.6                   & {100}         & {95.8}          & 87.3                   & 7.3 $\pm$ 1.6 & 0.190 $\pm$ 0.012 \\
BR-Net~\cite{adeli2021}                  & 94.8 $\pm$ 0.4 & 92.7                   & 97.4                   & 93.6                   & {100}         & {95.8}          & {{88.5}} & \textit{\textbf{5.3}} $\pm$ 1.2 & 0.170 $\pm$ 0.007 \\
Vanilla Diffusion Autoencoder           & 93.0 $\pm$ 0.6 & 85.4                   & 95.6                   & {100}         & 91.7                   & 87.3                   & {90.1}          & 11.5 $\pm$ 1.1 & 0.326 $\pm$ 0.000 \\
\midrule
$y$-space CI-DiffAE                             & \textit{\textbf{94.8}} $\pm$ 2.1 & 80.5 & 93.7         & 93.6 & 87.5                   & 81.1 & 86.2                   & 9.9 $\pm$ 6.0 & 0.120 $\pm$ 0.034  \\
$v$-space CI-DiffAE                             & \textbf{96.6} $\pm$ 1.8  & {{97.5}} & {100}         & {{97.9}} & 91.7                   & {{93.7}} & 87.3                   & \textbf{5.0} $\pm$ 2.5 & \textit{\textbf{0.076}} $\pm$ 0.025   \\
\bottomrule
\end{tabularx}
\end{table*}

In selecting an experimental setting in which to apply our architecture to the diffusion autoencoder, we searched for datasets (1) labeled with a relevant secondary attribute that we could treat as a confounding variable, and (2) compatible with other public datasets that could be used to pretrain the DiffAE before finetuning. One of the few datasets meeting these criteria was a set of 1,270 face images provided by the Gender Shades project \cite{buolamwini2018gender}, representing subjects from 3 African countries and 3 European countries. Each image is labeled with both skin type, using the Fitzpatrick classification system, and sex, inheriting a binary male/female grouping as a simplified proxy for gender. We aligned and cropped the images to the format expected by the diffusion autoencoder model, and then employed our architecture to train the diffusion autoencoder to predict sex while being conditionally independent to skin type.

We initialized a pretrained diffusion model based on the $128\times128$ Flickr-Faces-HQ Dataset, which consists of 70,000 highly varied face images. We treat the model's semantic subcode as the latent space and add a single linear layer prediction head to convert from this 512-length vector space to a single logit representing the sex label space.

\paragraph{Models B1-3} We implement three baseline models. BR-Net \cite{adeli2021} uses adversarial training to learn image features that have statistical ``mean independence"  with the protected attribute. The adversarial objective of BR-Net is based on Pearson's correlation between the true and the predicted value of the protected attribute. Second, \citeauthor{kim2019} uses mutual information minimization to ensure that the learned image features cannot predict the protected attribute. Finally, a multi-task network \cite{lu_multitask2017} is implemented to use the same encoder to predict both the target value and the protected attribute simultaneously.

\paragraph{Model B4: Vanilla Diffusion Autoencoder}
For a closer comparison to our architecture, we train the diffusion autoencoder using its default loss function, denoted in \cite{preechakul2022diffusion} as $L_\text{simple}$. In a separate optimization step, we train the prediction head using binary cross entropy loss (BCE).

\paragraph{Model 1: $y$-space CI-DiffAE}
We unify the training of the diffusion autoencoder and the prediction head under a single loss function and enforce conditional independence with respect to the output of the prediction head. In the context of \eqref{eqn:enc_loss}, the encoder objective function becomes
\begin{equation}
    L_\text{Enc} = \lambda L_{\text{CI}} + \left( L_{\text{DiffAE}} + \rho ~\text{BCE}(y, \hat y)\right),
\end{equation}
for hyperparameter $\rho$.

\begin{figure}
    \centering
    \includegraphics[width=\linewidth]{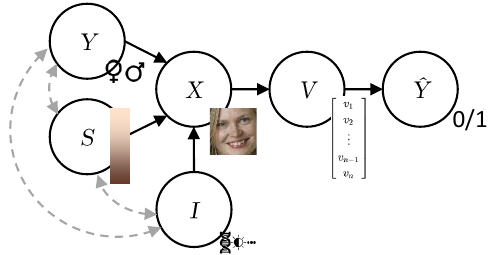}
    \vspace{-8pt}
    \caption{Causal graph corresponding to face dataset. By way of the latent space $V$, we train a model to predict sex ($Y$) from face image ($X$) while being invariant to skin color ($S$). Here, countless other factors ($I$) such as genetics and lighting also influence each image, and hidden confounders (dashed lines) cause correlation among inputs.}
    \label{fig:causalgraph}
\end{figure}

\begin{figure*}[t]
    \centering
    \includegraphics[width=\linewidth]{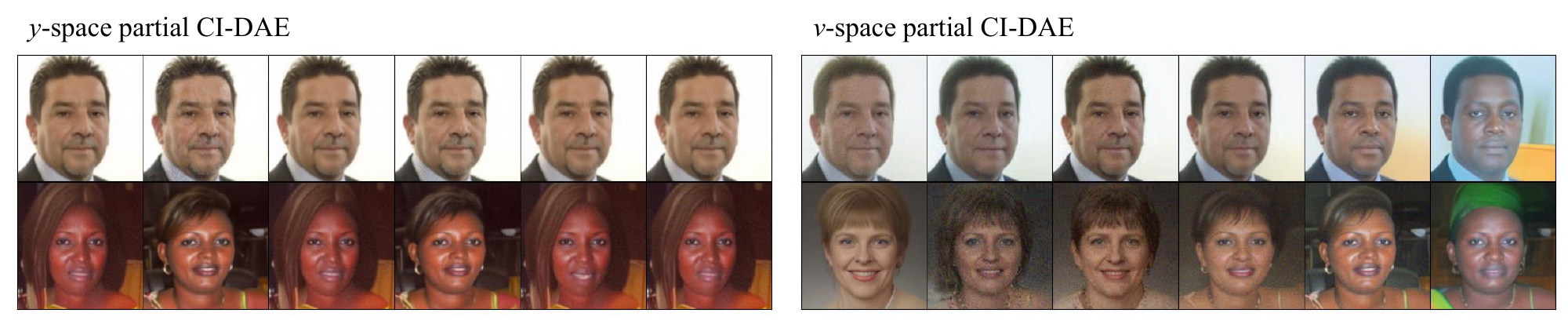} \vspace{-15pt}
    \caption{Selected image reconstructions. For each variant of our CI-DiffAE, we reconstruct two images from the dataset while adjusting the race-variant portion of the semantic subcode: the leftmost image in each group corresponds to skin type 1, and the rightmost to skin type 6. The latent  ($v$-space) CI constraint can effectively disentangle skin shade from other facial features. }\vspace{-3pt}
    \label{fig:dae_examples}
\end{figure*}

\begin{figure}[t]
    \centering
    \includegraphics[width=\linewidth]{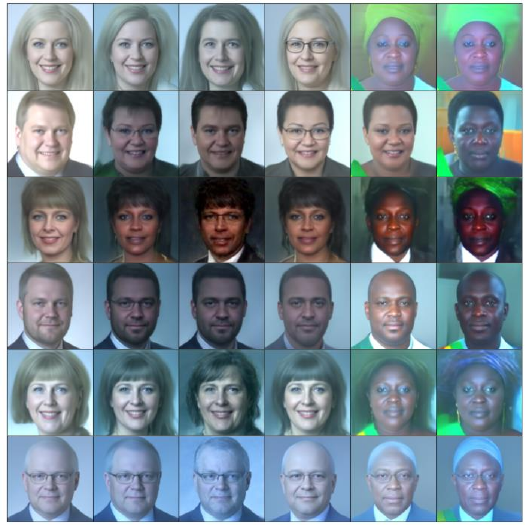} \vspace{-15pt}
    \caption{Race-invariant image generation using our $v$-space partial CI-DiffAE architecture. Each column contains hallucinated images of the same skin type.}\vspace{-3pt}
    \label{fig:dae_averages}
\end{figure}

\paragraph{Model 2: $v$-space CI-DiffAE}
In an analogous manner to the $v$-space CI-CNN, we enforce conditional independence directly on the semantic subcode, keeping other portions of the model the same.

\paragraph{}
Results in Table \ref{tab:face-results} indicate that the $v$-space constraint yields an all-around improvement in accuracy and fairness compared to the vanilla model, as opposed to the $y$-space constraint which achieves lower dcor$^2$ with considerable impact on accuracy. This improvement is statistically significant as measured by a McNemar test ($\chi^2=8.257$, $p\text{-value}=0.0040$). Overall, these results demonstrate that the $v$-space CI constraint produces results well on-par with existing work. Moreover, this technique has the added capability of causal image generation, described below.

\subsubsection{Visual Results}
Due to the strong association between skin type and race, skin type impacts a large number of facial features. Our architecture removes much of this association, which creates a race-invariant representation, but as a result, also constricts the model's ability to encode anatomical facial features in the semantic subcode. Therefore, nearly all information becomes encoded by the stochastic subcode, and generated images are less realistic.

As discussed previously, we remedy this effect by enforcing conditional independence on only a portion of the semantic subcode and simultaneously training a sensitive attribute prediction head on the remaining portion. We find that this partial CI-DiffAE architecture allows the semantic subcode as a whole to continue encoding the facial features necessary to produce a clean image. To interpret the effects on the semantic subcode, we generate images after adjusting the race-variant portion of the semantic subcode to values corresponding to each skin type (Fig. \ref{fig:dae_examples}). We observe that the $v$-space model is effectively (and causally) able to disentangle race-related features within the semantic subcode, resulting in a smooth and meaningful transition with changes limited to those relating to race.

Furthermore, by sampling the race-invariant portion of the semantic subcode, we retain the diffusion autoencoder's ability to generate novel images, but with added control over skin type (Fig. \ref{fig:dae_averages}). These images demonstrate how our technique creates the race-invariant portion of the latent representation by removing the direct causal effect of skin shade, analogous to deleting the solid arrow between $S$ and $X$ in Fig. \ref{fig:causalgraph}. At the same time, due to the fine-grained nature of the equalized odds criterion we enforce, the representation preserves the indirect effect of skin shade that is mediated by sex, shown in Fig. \ref{fig:causalgraph} as the dashed arrow between $Y$ and $S$. Therefore, the model continues to capture differences in gender expression across race, most notably in the headwraps often worn by women in Sub-Saharan African cultures. By disentangling race from the latent representation, we gain a richer insight into the causal processes that underlie facial images, crucial for effective image generation.
\section{Conclusion}

We introduced a framework to ensure fair and unconfounded representation learning during training and demonstrated both its versatility when applied to complex models and its effectiveness when compared to alternative methods. We iterated upon the theoretical idea of expressing the conditional independence constraint as an equality of two Jensen-Shannon divergences and extended this to high dimensional latent space via a dynamic sampling technique that can be easily implemented for any encoder. Our work exposes a new approach to generally enforce a conditional independence constraint on a model, which can then be used in downstream tasks such as causal image generation and fair predictive models. To our knowledge, this is the only model-agnostic training approach to be shown effective on enforcing specific features to be encoded in given dimensions of the latent space.
We are optimistic that further experimentation will reveal applications to other tasks concerning fairness and disentanglement. Moreover, as a direction for future work, the use of conditionally invariant embeddings may prove useful in extending traditional causal methods like mediational analysis to complex, high-dimensional settings.
\section*{Acknowledgement}
This research was partially supported by UST, Stanford Institute for Human-Centered AI (HAI) GCP cloud credits, and National Institutes of Health (NIH) grants U54HG012510 and AG084471. J.H. was also supported by a research fund from Panasonic and E.A. by the Stanford School of Medicine Jaswa Innovator Award. 
{
    \small
    \bibliographystyle{ieeenat_fullname}
    \bibliography{main}

\begin{thebibliography}{37}
\providecommand{\natexlab}[1]{#1}
\providecommand{\url}[1]{\texttt{#1}}
\expandafter\ifx\csname urlstyle\endcsname\relax
  \providecommand{\doi}[1]{doi: #1}\else
  \providecommand{\doi}{doi: \begingroup \urlstyle{rm}\Url}\fi

\bibitem[Adeli et~al.(2021)Adeli, Zhao, Pfefferbaum, Sullivan, Fei-Fei, Niebles, and Pohl]{adeli2021}
Ehsan Adeli, Qingyu Zhao, Adolf Pfefferbaum, Edith~V. Sullivan, Li Fei-Fei, Juan~Carlos Niebles, and Kilian~M. Pohl.
\newblock Representation learning with statistical independence to mitigate bias.
\newblock In \emph{2021 IEEE Winter Conference on Applications of Computer Vision (WACV)}, pages 2512--2522, 2021.

\bibitem[Agrawal et~al.(2021)Agrawal, Squires, Prasad, and Uhler]{agrawal2021}
Raj Agrawal, Chandler Squires, Neha Prasad, and Caroline Uhler.
\newblock The decamfounder: Non-linear causal discovery in the presence of hidden variables, 2021.

\bibitem[Ahuja et~al.(2021)Ahuja, Sattigeri, Shanmugam, Wei, Natesan~Ramamurthy, and Kocaoglu]{pmlr-v161-ahuja21a}
Kartik Ahuja, Prasanna Sattigeri, Karthikeyan Shanmugam, Dennis Wei, Karthikeyan Natesan~Ramamurthy, and Murat Kocaoglu.
\newblock Conditionally independent data generation.
\newblock In \emph{Proceedings of the Thirty-Seventh Conference on Uncertainty in Artificial Intelligence}, pages 2050--2060. PMLR, 2021.

\bibitem[Akuzawa et~al.(2019)Akuzawa, Iwasawa, and Matsuo]{akuzawa2019adversarial}
Kei Akuzawa, Yusuke Iwasawa, and Yutaka Matsuo.
\newblock Adversarial invariant feature learning with accuracy constraint for domain generalization.
\newblock \emph{arXiv preprint arXiv:1904.12543}, 2019.

\bibitem[Buolamwini and Gebru(2018)]{buolamwini2018gender}
Joy Buolamwini and Timnit Gebru.
\newblock Gender shades: Intersectional accuracy disparities in commercial gender classification.
\newblock In \emph{Conference on fairness, accountability and transparency}, pages 77--91, 2018.

\bibitem[Feldman(2015)]{feldman2015computational}
Michael Feldman.
\newblock \emph{Computational fairness: Preventing machine-learned discrimination}.
\newblock PhD thesis, Haverford College, 2015.

\bibitem[Giannone et~al.(2022)Giannone, Nielsen, and Winther]{giannone2022few}
Giorgio Giannone, Didrik Nielsen, and Ole Winther.
\newblock Few-shot diffusion models.
\newblock \emph{arXiv preprint arXiv:2205.15463}, 2022.

\bibitem[Goodfellow et~al.(2020)Goodfellow, Pouget-Abadie, Mirza, Xu, Warde-Farley, Ozair, Courville, and Bengio]{goodfellow2020generative}
Ian Goodfellow, Jean Pouget-Abadie, Mehdi Mirza, Bing Xu, David Warde-Farley, Sherjil Ozair, Aaron Courville, and Yoshua Bengio.
\newblock Generative adversarial networks.
\newblock \emph{Communications of the ACM}, 63\penalty0 (11):\penalty0 139--144, 2020.

\bibitem[Guan et~al.(2020)Guan, Zhang, and Lu]{guan2020large}
Jiechao Guan, Manli Zhang, and Zhiwu Lu.
\newblock Large-scale cross-domain few-shot learning.
\newblock In \emph{Proceedings of the Asian Conference on Computer Vision}, 2020.

\bibitem[Hardt et~al.(2016)Hardt, Price, and Srebro]{hardt2016equality}
Moritz Hardt, Eric Price, and Nati Srebro.
\newblock Equality of opportunity in supervised learning.
\newblock In \emph{Advances in neural information processing systems}, pages 3315--3323, 2016.

\bibitem[Islam et~al.(2022)Islam, Fariha, Meliou, and Salimi]{islam2022through}
Maliha~Tashfia Islam, Anna Fariha, Alexandra Meliou, and Babak Salimi.
\newblock Through the data management lens: Experimental analysis and evaluation of fair classification.
\newblock In \emph{Proceedings of the 2022 International Conference on Management of Data}, pages 232--246, 2022.

\bibitem[Johndrow et~al.(2019)Johndrow, Lum, et~al.]{johndrow2019algorithm}
James~E Johndrow, Kristian Lum, et~al.
\newblock An algorithm for removing sensitive information: application to race-independent recidivism prediction.
\newblock \emph{The Annals of Applied Statistics}, 13\penalty0 (1):\penalty0 189--220, 2019.

\bibitem[Kim et~al.(2019)Kim, Kim, Kim, Kim, and Kim]{kim2019}
Byungju Kim, Hyunwoo Kim, Kyungsu Kim, Sungjin Kim, and Junmo Kim.
\newblock Learning not to learn: Training deep neural networks with biased data.
\newblock In \emph{Proceedings of the IEEE Conference on Computer Vision and Pattern Recognition}, pages 9012--9020, 2019.

\bibitem[Kusner et~al.(2017)Kusner, Loftus, Russell, and Silva]{kusner2017counterfactual}
Matt~J Kusner, Joshua Loftus, Chris Russell, and Ricardo Silva.
\newblock Counterfactual fairness.
\newblock \emph{Advances in neural information processing systems}, 30, 2017.

\bibitem[Li et~al.(2018)Li, Pan, Wang, and Kot]{li2018domain}
Haoliang Li, Sinno~Jialin Pan, Shiqi Wang, and Alex~C Kot.
\newblock Domain generalization with adversarial feature learning.
\newblock In \emph{Proceedings of the IEEE conference on computer vision and pattern recognition}, pages 5400--5409, 2018.

\bibitem[Lu et~al.(2021)Lu, Zhao, Zhang, Pohl, Fei-Fei, Niebles, and Adeli]{lu2021metadata}
Mandy Lu, Qingyu Zhao, Jiequan Zhang, Kilian~M Pohl, Li Fei-Fei, Juan~Carlos Niebles, and Ehsan Adeli.
\newblock Metadata normalization.
\newblock In \emph{Proceedings of the IEEE/CVF Conference on Computer Vision and Pattern Recognition}, pages 10917--10927, 2021.

\bibitem[Lu et~al.(2017)Lu, Kumar, Zhai, Cheng, Javidi, and Feris]{lu_multitask2017}
Yongxi Lu, Abhishek Kumar, Shuangfei Zhai, Yu Cheng, Tara Javidi, and Rogerio Feris.
\newblock Fully-adaptive feature sharing in multi-task networks with applications in person attribute classification.
\newblock In \emph{2017 IEEE Conference on Computer Vision and Pattern Recognition (CVPR)}, pages 1131--1140, 2017.

\bibitem[Madras et~al.(2018)Madras, Creager, Pitassi, and Zemel]{madras2018learning}
David Madras, Elliot Creager, Toniann Pitassi, and Richard Zemel.
\newblock Learning adversarially fair and transferable representations.
\newblock \emph{arXiv preprint arXiv:1802.06309}, 2018.

\bibitem[McNamee(2005)]{mcnamee2005regression}
Roseanne McNamee.
\newblock Regression modelling and other methods to control confounding.
\newblock \emph{Occupational and environmental medicine}, 62\penalty0 (7):\penalty0 500--506, 2005.

\bibitem[Mehrabi et~al.(2021)Mehrabi, Morstatter, Saxena, Lerman, and Galstyan]{mehrabi2021survey}
Ninareh Mehrabi, Fred Morstatter, Nripsuta Saxena, Kristina Lerman, and Aram Galstyan.
\newblock A survey on bias and fairness in machine learning.
\newblock \emph{ACM Computing Surveys (CSUR)}, 54\penalty0 (6):\penalty0 1--35, 2021.

\bibitem[Moyer et~al.(2018)Moyer, Gao, Brekelmans, Galstyan, and Ver~Steeg]{moyer2018}
Daniel Moyer, Shuyang Gao, Rob Brekelmans, Aram Galstyan, and Greg Ver~Steeg.
\newblock Invariant representations without adversarial training.
\newblock In \emph{Advances in Neural Information Processing Systems 31}, pages 9084--9093. Curran Associates, Inc., 2018.

\bibitem[Pirhadi et~al.(2024)Pirhadi, Moslemi, Cloninger, Milani, and Salimi]{pirhadi2024otclean}
Alireza Pirhadi, Mohammad~Hossein Moslemi, Alexander Cloninger, Mostafa Milani, and Babak Salimi.
\newblock Otclean: Data cleaning for conditional independence violations using optimal transport.
\newblock \emph{SIGMOD}, 2024.

\bibitem[Preechakul et~al.(2022)Preechakul, Chatthee, Wizadwongsa, and Suwajanakorn]{preechakul2022diffusion}
Konpat Preechakul, Nattanat Chatthee, Suttisak Wizadwongsa, and Supasorn Suwajanakorn.
\newblock Diffusion autoencoders: Toward a meaningful and decodable representation.
\newblock In \emph{Proceedings of the IEEE/CVF Conference on Computer Vision and Pattern Recognition}, pages 10619--10629, 2022.

\bibitem[Salimi et~al.(2019)Salimi, Rodriguez, Howe, and Suciu]{salimi2019interventional}
Babak Salimi, Luke Rodriguez, Bill Howe, and Dan Suciu.
\newblock Interventional fairness: Causal database repair for algorithmic fairness.
\newblock In \emph{Proceedings of the 2019 International Conference on Management of Data}, pages 793--810, 2019.

\bibitem[Salimi et~al.(2020)Salimi, Howe, and Suciu]{salimi2020}
Babak Salimi, Bill Howe, and Dan Suciu.
\newblock Database repair meets algorithmic fairness.
\newblock \emph{ACM SIGMOD Record}, 2020.

\bibitem[Sinha et~al.(2021)Sinha, Song, Meng, and Ermon]{sinha2021d2c}
Abhishek Sinha, Jiaming Song, Chenlin Meng, and Stefano Ermon.
\newblock D2c: Diffusion-decoding models for few-shot conditional generation.
\newblock \emph{Advances in Neural Information Processing Systems}, 34:\penalty0 12533--12548, 2021.

\bibitem[Squires et~al.(2020)Squires, Shen, Agarwal, Shah, and Uhler]{squires2020}
Chandler Squires, Dennis Shen, Anish Agarwal, Devavrat Shah, and Caroline Uhler.
\newblock Causal imputation via synthetic interventions, 2020.

\bibitem[Sz{\'e}kely et~al.(2007)Sz{\'e}kely, Rizzo, Bakirov, et~al.]{szekely2007measuring}
G{\'a}bor~J Sz{\'e}kely, Maria~L Rizzo, Nail~K Bakirov, et~al.
\newblock Measuring and testing dependence by correlation of distances.
\newblock \emph{The annals of statistics}, 35\penalty0 (6):\penalty0 2769--2794, 2007.

\bibitem[Tan et~al.(2020)Tan, Yeom, Fredrikson, and Talwalkar]{tan2020learning}
Zilong Tan, Samuel Yeom, Matt Fredrikson, and Ameet Talwalkar.
\newblock Learning fair representations for kernel models.
\newblock In \emph{International Conference on Artificial Intelligence and Statistics}, pages 155--166. PMLR, 2020.

\bibitem[Vento et~al.(2022)Vento, Zhao, Paul, Pohl, and Adeli]{vento2022penalty}
Anthony Vento, Qingyu Zhao, Robert Paul, Kilian~M Pohl, and Ehsan Adeli.
\newblock A penalty approach for normalizing feature distributions to build confounder-free models.
\newblock In \emph{International Conference on Medical Image Computing and Computer-Assisted Intervention}, pages 387--397. Springer, 2022.

\bibitem[Wang et~al.(2019)Wang, Zhao, Yatskar, Chang, and Ordonez]{wang2019balanced}
Tianlu Wang, Jieyu Zhao, Mark Yatskar, Kai-Wei Chang, and Vicente Ordonez.
\newblock Balanced datasets are not enough: Estimating and mitigating gender bias in deep image representations.
\newblock In \emph{Proceedings of the IEEE/CVF International Conference on Computer Vision}, pages 5310--5319, 2019.

\bibitem[Woodworth et~al.(2017)Woodworth, Gunasekar, Ohannessian, and Srebro]{woodworth2017learning}
Blake Woodworth, Suriya Gunasekar, Mesrob~I Ohannessian, and Nathan Srebro.
\newblock Learning non-discriminatory predictors.
\newblock \emph{arXiv preprint arXiv:1702.06081}, 2017.

\bibitem[Zhang et~al.(2018)Zhang, Lemoine, and Mitchell]{zhang2018mitigating}
Brian~Hu Zhang, Blake Lemoine, and Margaret Mitchell.
\newblock Mitigating unwanted biases with adversarial learning.
\newblock In \emph{Proceedings of the 2018 AAAI/ACM Conference on AI, Ethics, and Society}, pages 335--340, 2018.

\bibitem[Zhang et~al.(2022)Zhang, Shao, Ma, Lv, and Zhou]{zhang2022enhanced}
Zhongwei Zhang, Mingyu Shao, Chicheng Ma, Zhe Lv, and Jilei Zhou.
\newblock An enhanced domain-adversarial neural networks for intelligent cross-domain fault diagnosis of rotating machinery.
\newblock \emph{Nonlinear Dynamics}, 108\penalty0 (3):\penalty0 2385--2404, 2022.

\bibitem[Zhao et~al.(2019{\natexlab{a}})Zhao, Coston, Adel, and Gordon]{zhao2019conditional}
Han Zhao, Amanda Coston, Tameem Adel, and Geoffrey~J Gordon.
\newblock Conditional learning of fair representations.
\newblock \emph{arXiv preprint arXiv:1910.07162}, 2019{\natexlab{a}}.

\bibitem[Zhao et~al.(2019{\natexlab{b}})Zhao, Adeli, Pfefferbaum, Sullivan, and Pohl]{zhao2019confounder}
Qingyu Zhao, Ehsan Adeli, Adolf Pfefferbaum, Edith~V Sullivan, and Kilian~M Pohl.
\newblock Confounder-aware visualization of convnets.
\newblock \emph{arXiv preprint arXiv:1907.12727}, 2019{\natexlab{b}}.

\bibitem[Zhao et~al.(2020)Zhao, Adeli, and Pohl]{zhao2020training}
Qingyu Zhao, Ehsan Adeli, and Kilian~M Pohl.
\newblock Training confounder-free deep learning models for medical applications.
\newblock \emph{Nature communications}, 11\penalty0 (1):\penalty0 1--9, 2020.

\end{thebibliography}
}
\end{document}